\documentclass[letterpaper, 10 pt, journal, twoside]{IEEEtran}  

\IEEEoverridecommandlockouts                              

\usepackage{amssymb,amsfonts,amsmath,amscd}
\usepackage{cite}
\usepackage{graphicx}
\usepackage{subfigure}
\usepackage{hyperref}

\usepackage{tikz}
\usetikzlibrary{calc}
\usetikzlibrary{arrows}
\usepackage{pgfplots}

\newcommand{\bcf}{\;\mbox{\boldmath ${\cal F}$\unboldmath}}
\def\Vec#1{\!\!\hbox{$#1$\kern-0.38em\lower0.85em\hbox{$\vec{}\,$}}\,}%
\newcommand{\bbm}{\begin{bmatrix}}
\newcommand{\ebm}{\end{bmatrix}}
\DeclareMathAlphabet{\mbf}{OT1}{ptm}{b}{n}
\newcommand{\mbs}[1]{{\boldsymbol{#1}}}

\newcommand{\pftm}{\odot}

\newcommand{\change}[1]{{\color{black} #1}}

\title{Do We Need to Compensate for Motion Distortion and Doppler Effects in Spinning Radar Navigation?
}

\author{Keenan Burnett, Angela P. Schoellig, Timothy D. Barfoot
\thanks{Manuscript received: October, 15, 2020; Revised December, 16, 2020; Accepted January, 13, 2021. This paper was recommended for publication by Associate Editor Javier Civera and Editor Allison Okamura upon evaluation of the Reviewers' comments.}
\thanks{All authors are affiliated with the University of Toronto Institute for Aerospace Studies (UTIAS):
        {\tt\scriptsize keenan.burnett@utoronto.ca}, {\tt\scriptsize schoellig@utias.utoronto.ca}, {\tt\scriptsize tim.barfoot@utoronto.ca}}%
}

\begin{document}
	
\maketitle

\bibliographystyle{IEEEtran}

\markboth{IEEE Robotics and Automation Letters. Preprint Version. Accepted January, 2021}
{Burnett \MakeLowercase{\textit{et al.}}: Do We Need to Compensate for Motion Distortion and Doppler Effects in Spinning Radar Navigation?} 

\begin{abstract}

In order to tackle the challenge of unfavorable weather conditions such as rain and snow, radar is being revisited as a parallel sensing modality to vision and lidar. Recent works have made tremendous progress in applying spinning radar to odometry and place recognition. However, these works have so far ignored the impact of motion distortion and Doppler effects on spinning-radar-based navigation, which may be significant in the self-driving car domain where speeds can be high. In this work, we demonstrate the effect of these distortions on radar odometry using the Oxford Radar RobotCar Dataset and metric localization using our own data-taking platform.  We revisit a lightweight estimator that can recover the motion between a pair of radar scans while accounting for both effects. Our conclusion is that both motion distortion and the Doppler effect are significant in different aspects of spinning radar navigation, with the former more prominent than the latter. Code for this project can be found at: {\scriptsize \url{https://github.com/keenan-burnett/yeti_radar_odometry}}






\end{abstract}


\vspace{-1mm}

\section{INTRODUCTION}

\vspace{-1mm}

\IEEEPARstart{A}{s} researchers continue to advance the capabilities of autonomous vehicles, attention has begun to shift towards inclement weather conditions. Currently, most autonomous vehicles rely primarily cameras and lidar for perception and localization. Although these sensors have been shown to achieve sufficient performance under nominal conditions, rain and snow remain an open problem. Radar sensors, such as the one produced by Navtech \cite{navtech}, may provide a solution.

\vspace{-0.5mm}


Due to its longer wavelength, radar is robust to small particles such as dust, fog, rain, or snow, which can negatively impact cameras and lidar sensors. Furthermore, radar sensors tend to have a longer detection range and can penetrate through some materials allowing them to see beyond the line of sight of lidar. These features make radar particularly well-suited for inclement weather. However, radar sensors have coarser spatial resolution than lidar and suffer from a higher noise floor making them challenging to work with.


Recent works have made tremendous progress in applying the Navtech radar to odometry \cite{cen_icra18, cen_icra19, aldera_icra19, barnes_corl19, barnes_icra20, park_icra20} and place recognition \cite{saftescu_icra20, gadd_arxiv20, kim_icra20}. However, all of these works make the simplifying assumption that a radar scan is collected at a single instant in time. In reality, the sensor is rotating while the vehicle is moving causing the radar scan to be distorted in a cork-screw fashion. Range measurements of the Navtech radar are also impacted by Doppler frequency shifts resulting from the relative velocity between the sensor and its surroundings. Both distortion effects become more pronounced as the speed of the ego-vehicle increases. \change{Most automotive radar sensors are not impacted by either distortion effect. However, the Navtech provides $360^\circ$ coverage with accurate range and azimuth resolution, making it an appealing navigation sensor.}

\vspace{-0.5mm}


In this paper, we demonstrate the effect that motion distortion can have on radar-based navigation. We also \change{revisit} a lightweight estimator, Motion-Compensated RANSAC \cite{anderson_iros13}, which can recover the motion between a pair of scans and remove the distortion. The Doppler effect was briefly acknowledged in \cite{cen_icra18} but our work is the first to demonstrate the impact on radar-based navigation and to provide a method for its compensation.

\vspace{-0.5mm}





\begin{figure}[t]
	\includegraphics[width=0.95\columnwidth]{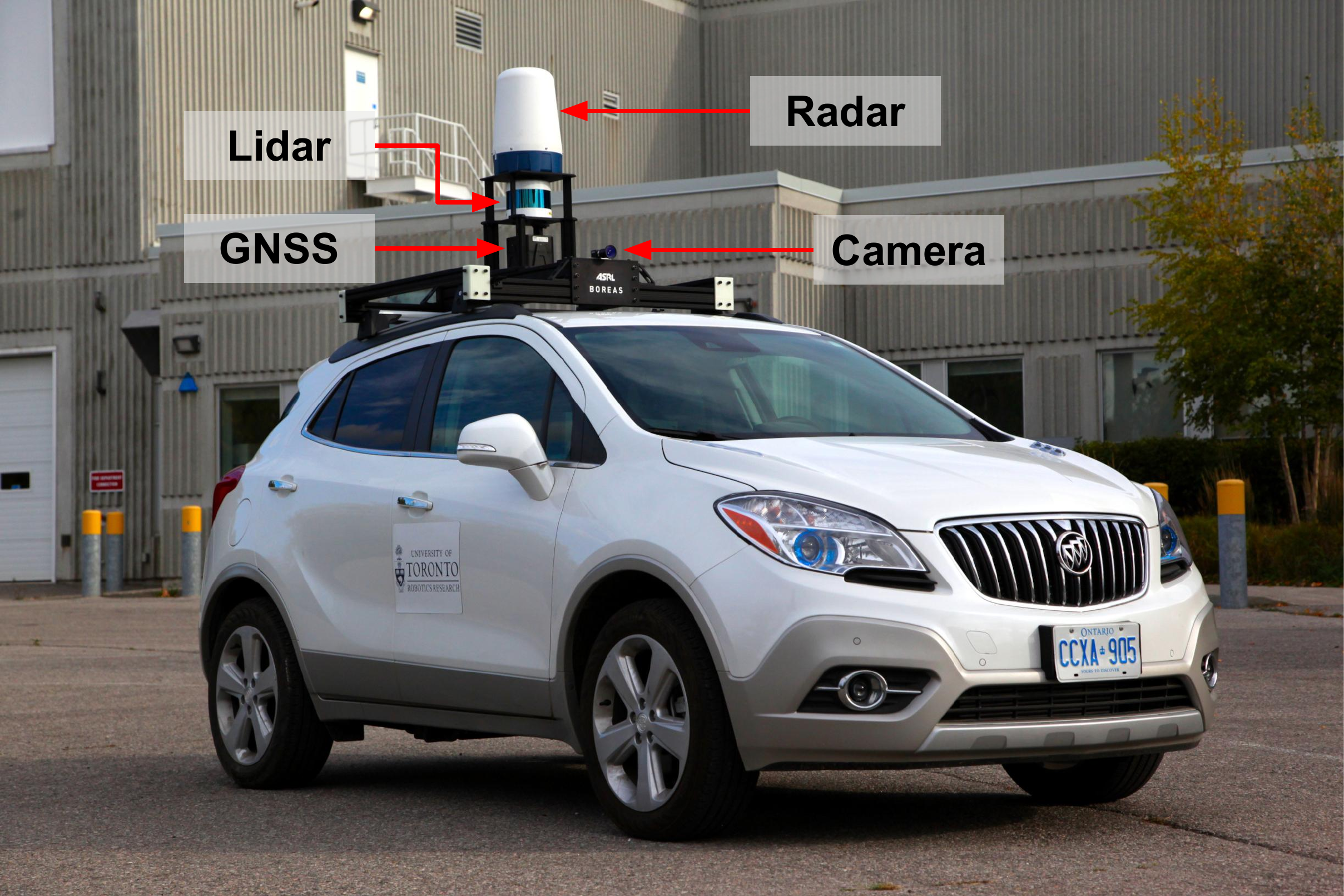}
	\centering
	\caption{Our data-taking platform, \textit{Boreas}, which includes a Velodyne Alpha-Prime (128-beam) lidar, Navtech CIR204-H radar, FLIR Blackfly~S monocular camera, and Applanix POSLV GNSS.}
	\label{fig:buick}
	\vspace{-7mm}
\end{figure}

As our primary experiment to demonstrate the effects of motion distortion, we perform radar odometry on the Oxford Radar RobotCar Dataset \cite{oxford_radar}. As an additional experiment, we perform metric localization using our own data-taking platform, shown in Figure~\ref{fig:buick}. \change{Qualitative results of both distortion effects are also provided.} Rather than focusing on achieving state-of-the-art navigation results, the goal of this paper is to show that motion distortion and Doppler effects are significant and can be compensated for with relative ease.

\vspace{-0.5mm}

The rest of this paper is organized as follows:  Sec.\ II discusses related work, III provides our methodology to match two radar scans while compensating for motion distortion and the Doppler effect, IV has experiments, and V concludes.

\vspace{-1.5mm}





\section{Related Work}

\vspace{-1mm}

Self-driving vehicles designed to operate in ideal conditions often relegate radar to a role as a secondary sensor as part of an emergency braking system \cite{burnett_jfr20}. However, recent advances in Frequency Modulated Continuous Wave (FMCW) radar indicate that it is a promising sensor for navigation and other tasks typically reserved for vision and lidar. Jose and Adams \cite{jose_iros04} \cite{jose_sens10} were the first to research the application of spinning radar to outdoor SLAM. In \cite{checchin_fsr10}, Checchin et al. present one of the first mapping and localization systems designed for a FMCW scanning radar. Their approach finds the transformation between pairs of scans using the Fourier Mellin transform. Vivet et al. \cite{vivet_sens13} were the first to address the motion distortion problem for scanning radar. Our approach to motion distortion is simpler and we offer a more up-to-date analysis on a dataset that is relevant to current research. In \cite{kellner_itsc13}, Kellner et al. present a method that estimates the linear and angular velocity of the ego-vehicle when the Doppler velocity of each target is known. \change{The use of single-chip FMCW radar has recently become a salient area of research for robust indoor positioning systems under conditions unfavorable to vision \cite{kramer_icra20, lu_censs20}.}


\vspace{-0.5mm}


\begin{figure}[t]
	\centering
	\subfigure[Feature Extraction]{\includegraphics[width=0.47\columnwidth]{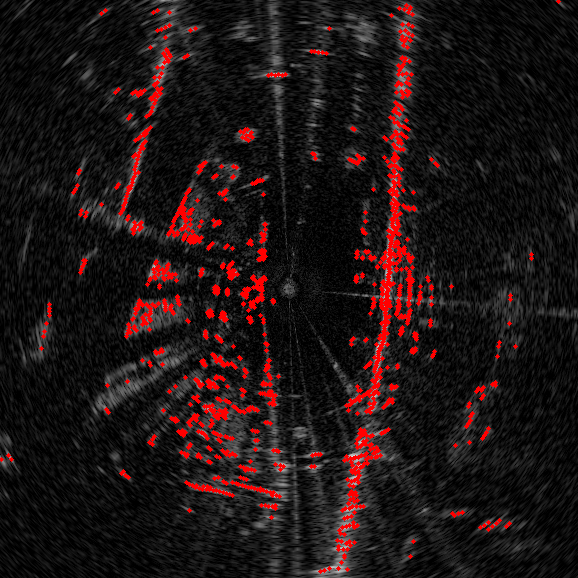}}
	\subfigure[Data Association]{\includegraphics[width=0.47\columnwidth]{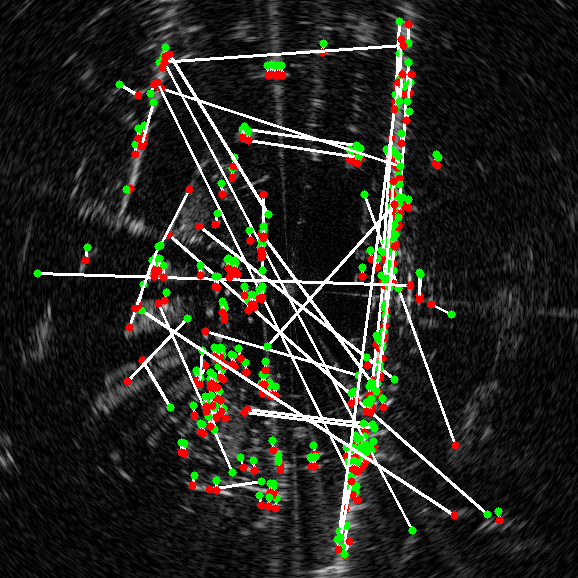}}
	\subfigure[MC-RANSAC Inliers]{\includegraphics[width=0.47\columnwidth]{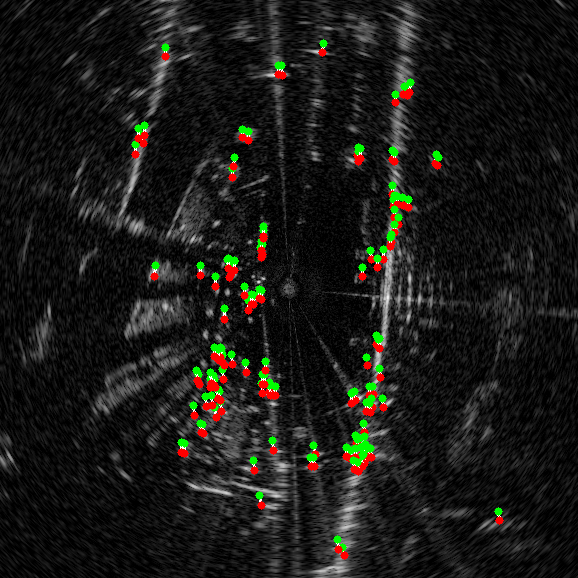}}
	\subfigure[MC-RANSAC Inliers Rotating]{\includegraphics[width=0.47\columnwidth]{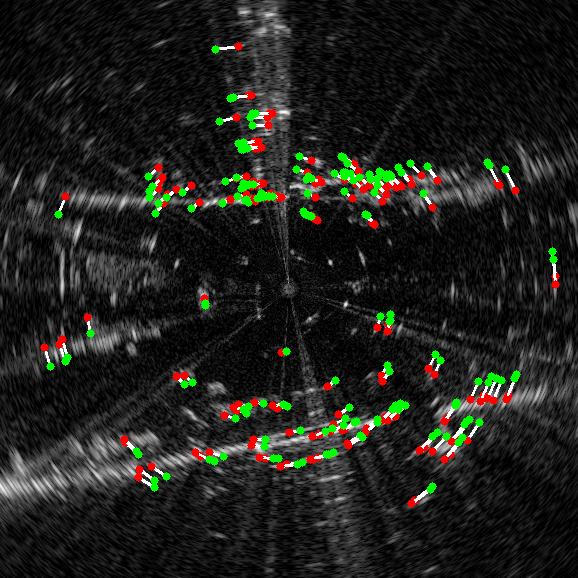}}
	\caption{This figure illustrates our feature extraction and matching process. (a) displays our raw extracted features. (b) displays the output of our ORB-descriptor-based data association. (c) and (d) show the inlier set resulting from motion-compensated RANSAC while driving straight and rotating.}
	\label{fig:features}
	\vspace{-8mm}
\end{figure}

In \cite{cen_icra18}, Cen et al. present their seminal work that has rekindled interest in applying FCMW radar to navigation. Their work presented a new method to extract stable keypoints and perform scan matching using graph matching. Further research in this area has been spurred by the introduction of the Oxford Radar RobotCar Dataset \cite{oxford_radar}, which includes lidar, vision, and radar data from a Navtech radar. \change{Other radar-focused datasets include the MulRan place recognition dataset \cite{kim_icra20} and the RADIATE object detection dataset \cite{sheeny_arxiv20}.}

\vspace{-0.5mm}

Odometry has recently been a central focus of radar-based navigation research. Components of an odometry pipeline can be repurposed for mapping and localization, which is an ultimate goal of this research. In \cite{cen_icra19}, Cen et al. present an update to their radar odometry pipeline with improved keypoint detection, descriptors, and a new graph matching strategy. Aldera et al. \cite{aldera_icra19} train a focus of attention policy to downsample the measurements given to data association, thus speeding up the odometry pipeline. In \cite{barnes_icra20}, Barnes and Posner present a deep-learning-based keypoint detector and descriptor that are learned directly from radar data using differentiable point matching and pose estimation. \change{In \cite{hong_iros20}, Hong et al. demonstrate the first radar-SLAM pipeline capable of handling extreme weather.}
\vspace{-0.5mm}

Other approaches forego the feature extraction process and instead use the entire radar scan for correlative scan matching.  Park et al. \cite{park_icra20} use the Fourier Mellin Transform on Cartesian and log-polar radar images to sequentially estimate rotation and translation. In \cite{barnes_corl19}, Barnes et al. present a fully differentiable, correlation-based radar odometry approach. In their system, a binary mask is learned such that unwanted distractor features are ignored by the scan matching. This approach currently represents the state of the art for radar odometry performance.
\vspace{-0.5mm}

Still others focus on topological localization that can be used by downstream metric mapping and localization systems to identify loop closures. S{\u{a}}ftescu et al. \cite{saftescu_icra20} learn a metric space embedding for radar scans using a convolutional neural network. Nearest-neighbour matching is then used to recognize locations at test time. Gadd et al. \cite{gadd_arxiv20} improve this place recognition performance by integrating a rotationally invariant metric space embedding into a sequence-based trajectory matching system previously applied to vision. In \cite{tang_icra20}, Tang et al. focus on localization between a radar on the ground and overhead satellite imagery.

\vspace{-0.5mm}

Other recent works using \change{a spinning} radar include the work by Weston et al. \cite{weston_icra19} that learns to generate occupancy grids from raw radar scans by using lidar data as ground truth. Kaul et al. \cite{kaul_arxiv20} train a semantic segmentation model for radar data using labels derived from lidar- and vision-based semantic segmentation.


\vspace{-0.5mm}

Motion distortion has been treated in the literature through the use of continuous-time trajectory estimation \cite{anderson_iros13, barfoot_rss14, anderson_ar15, anderson_iros15, furgale_ijrr15} for lidars \cite{bosse2008map} and rolling-shutter cameras \cite{hedborg2012rolling}, but these tools are yet to be applied to spinning radar. \change{TEASER \cite{yang_itr20} demonstrates a robust registration algorithm with optimality guarantees. However, it assumes that pointclouds are collected at single instances in time. Thus, TEASER is impacted by motion distortion in the same manner as rigid RANSAC.}

\vspace{-0.5mm}

Our work focuses on the problem of motion distortion and Doppler effects using the Navtech radar sensor which has not received attention from these prior works. Ideally, our findings will inform future research in this area looking to advance the state of the art in radar-based navigation.

\vspace{-2mm}











\section{Methodology} \label{sec:method}

\vspace{-1mm}


Section~\ref{feat_extract} describes our approach to feature extraction and data association. In section~\ref{mot_distort}, we \change{describe a} motion-compensated estimator and a rigid estimator for comparison. Section~\ref{dopp_correct} explains how the Doppler effect impacts radar range measurements and how to compensate for it.


\vspace{-1.5mm}

\subsection{Feature Extraction} \label{feat_extract}

Feature detection in radar data is more challenging than in lidar or vision due to its higher noise floor and lower spatial resolution. Constant False Alarm Rate (CFAR) \cite{rohling_irs11} is a simple feature detector that is popular for use with radar. CFAR is designed to estimate the local noise floor and capture relative peaks in the radar data. One-dimensional CFAR can be applied to Navtech data by convolving each azimuth with a sliding-window detector. 


As discussed in \cite{cen_icra18}, CFAR is not the best detector for radar-based navigation. CFAR produces many redundant keypoints, is difficult to tune, and produces false positives due to the noise artifacts present in radar. Instead, Cen et al. \cite{cen_icra18} proposed a detector that estimates a signal's noise statistics and then scales the power at each range by the probability that it is a real detection. In \cite{cen_icra19}, Cen et al. proposed an alternative detector that identifies continuous regions of the scan with high intensity and low gradients. Keypoints are then extracted by locating the middle of each continuous region. We will refer to these detectors as Cen2018 and Cen2019, respectively.

The original formulations of these detectors did not lend themselves to real-time operation. As such, we made several modifications to improve the runtime. For Cen2018, we use a Gaussian filter instead of a binomial filter, and we calculate the mean of each azimuth instead of using a median filter. We do not remove multipath reflections.



Cen2019 was designed to be easier to tune and have less redundant keypoints. However, we found that by adjusting the probability threshold of detections, Cen2018 obtained better odometry performance when combined with our RANSAC-based scan matching. Based on these preliminary tests, we concluded that Cen2018 was the best choice for our experiments. Figure~\ref{fig:features}(a) shows Cen2018 features plotted on top of a Cartesian radar image.





We convert the raw radar scans output by the Navtech sensor, which are in polar form, into Cartesian form. We then calculate an ORB descriptor \cite{rublee2011orb} for each keypoint on the Cartesian image. There may be better keypoint descriptors for radar data, such as the learned feature descriptors employed in \cite{barnes_icra20}. However, ORB descriptors are quick to implement, rotationally invariant, and resistant to noise.


For data association, we perform brute-force matching of ORB descriptors. We then apply a nearest-neighbor distance ratio test \cite{lowe2004distinctive} in order to remove false matches. The remaining matches are sent to our RANSAC-based estimators. Figure~\ref{fig:features}(b) shows the result of the initial data association. Note that there are several outliers. Figure~\ref{fig:features}(c),(d) shows the remaining inliers after performing RANSAC.
\vspace{-1mm}





\subsection{Motion Distortion} \label{mot_distort}


The output of data association is not perfect and often contains outliers. As a result, it is common to employ an additional outlier rejection scheme during estimation. In this paper, we use RANSAC \cite{ransac} to find an outlier-free set that can then be used to estimate the desired transform. If we assume that two radar scans are taken at times $\bar{t}_1$ and $\bar{t}_2$, then the  transformation between them can be estimated directly using the approach described in \cite{arun1987least}. 




%
During each iteration of RANSAC, a random subset of size $S$ is drawn from the initial matches and a candidate transform is generated. If the number of inliers exceeds a desired threshold or a maximum number of iterations is reached, the algorithm terminates. Radar scans are 2D and as such we use $S = 2$. The estimation process is repeated on the largest inlier set to obtain a more accurate transform. We will refer to this approach as \textit{rigid} RANSAC.


Our derivation of motion-compensated RANSAC follows closely from \cite{anderson_iros13}. However, we are applying the algorithm to a scanning radar in 2D instead of a two-axis scanning lidar in 3D. Furthermore, our derivation is shorter and uses updated notation from \cite{barfoot2017state}.

The principal idea behind motion-compensated RANSAC is to estimate the velocity of the sensor instead of estimating a transformation. We make the simplifying assumption that the linear and angular velocity between a pair of scans is constant. The combined velocity vector $\mbs{\varpi}$ is defined as

\vspace{-1.5mm}

\begin{equation}
	\mbs{\varpi} = \bbm \mbs{\nu} \\ \mbs{\omega} \ebm,
\end{equation}
where $\mbs{\nu}$ and $\mbs{\omega}$ are the linear and angular velocity in the sensor frame. To account for motion distortion, we remove the assumption that radar scans are taken at a single instant in time. Data association produces two sets of corresponding measurements, $\mbf{y}_{m,1}$ and $\mbf{y}_{m,2}$, where $m = 1 ... M$. Each pair of features, $m$, is extracted from sequential radar frames 1 and 2 at times $t_{m,1}$ and $t_{m,2}$. The temporal difference between a pair of measurements is $\Delta t_m := t_{m,2} - t_{m,1}$. The generative model for measurements is given as 
 \vspace{-1.5mm}
\begin{align}
	\mbf{y}_{m,1} &:= \mbf{f}(\mbf{T}_s(t_{m,1})\mbf{p}_m) + \mbf{n}_{m,1}, \\
	\mbf{y}_{m,2} &:= \mbf{f}(\mbf{T}_s(t_{m,2})\mbf{p}_m) + \mbf{n}_{m,2}, \nonumber
\end{align}
where $\mbf{f}(\cdot)$ is a nonlinear transformation from Cartesian to cylindrical coordinates and $\mbf{T}_s(t)$ is a 4 x 4 homogeneous transformation matrix representing the pose of the sensor frame $\Vec{\bcf}_s$ with respect to the inertial frame $\Vec{\bcf}_i$ at time $t$. $\mbf{p}_m$ is the original landmark location in the inertial frame.  We assume that each measurement is corrupted by zero-mean Gaussian noise: $\mbf{n}_{m,1} \sim \mathcal{N}(\mbf{0}, \mbf{R}_{m,1})$. The transformation between a pair of measurements is defined as
\vspace{-1mm}
\begin{align}
	\mbf{T}_m := \mbf{T}_s(t_{m,2}) \mbf{T}_s(t_{m,1})^{-1}.
\end{align}
\vspace{-0.5mm}To obtain our objective function, we convert feature locations from polar coordinates into local Cartesian coordinates:
\begin{align}
	\mbf{p}_{m,2} &= \mbf{f}^{-1}(\mbf{y}_{m,2}), \\
	\mbf{p}_{m,1} &= \mbf{f}^{-1}(\mbf{y}_{m,1}).
\end{align}
\vspace{-0.5mm}We then use the local transformation $\mbf{T}_m$ to create a pseudomeasurement $\hat{\mbf{p}}_{m,2}$,
\vspace{-1mm}
\begin{align}
	\hat{\mbf{p}}_{m,2} &= \mbf{T}_m \mbf{p}_{m,1}.
\end{align}
\vspace{-1mm}The error is then defined in the sensor coordinates and summed over each pair of measurements to obtain our objective function:
\vspace{-1mm}
\begin{align}
	\mbf{e}_m &= \mbf{p}_{m,2} - \hat{\mbf{p}}_{m,2}, \label{error1} \\
	J(\mbs{\varpi}) &:= \frac{1}{2} \sum_{m=1}^{M} \mbf{e}_m^T \mbf{R}_{{\rm cart},m}^{-1} \mbf{e}_m. \label{objective}
\end{align}

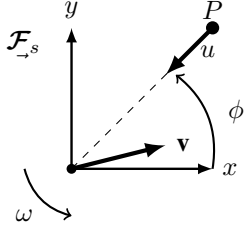
\begin{figure} [t]
	\centering
	
	\begin{tikzpicture} [axis/.style={>=latex,line width=0.9pt},
		vector/.style={line width=0.8pt},
		scale=1.25]
		\coordinate (I) at (0, 0);
		
		\draw [->, axis] (I) -- ($ (I) + (0, 1.5) $) node [above] {$y$};
		\draw [->, axis] (I) -- ($ (I) + (1.5, 0) $) node [right] {$x$} ;
		\fill[black] (I) circle (1.5pt);
		\draw ($ (I) + (-0.5, 1.25) $) node {$\Vec{\bcf}_s$};
		
		\coordinate (P) at ($ (I) + (1.5, 1.5) $);
		\fill[black] (P) circle (2.0pt) node (p) [above] {$P$};
		
		\draw[dashed] (I) -- (P);
		
		\path [->, line width=0.8pt, bend right] (1.5, 0) edge (1.11, 1.00);
		\draw node at (1.75, 0.61) {$\phi$};
		
		\draw [->, >=latex,line width=1.6pt] (I) -- (1.0, 0.25) node [right] {$\mbf{v}$};
		
		\path [->, line width=0.8pt, bend right] (-0.5, 0) edge (0, -0.5);
		\draw node at (-0.5, -0.5) {$\omega$};
		
		\draw [->, >=latex,line width=1.6pt] (P) -- ($ (P) - (0.5, 0.5) $) node [midway, right] {$u$};
		
	\end{tikzpicture}
	\caption{This diagram illustrates the relationship between the ego-motion ($\mbf{v}, \omega$) and the radial velocity $u$.}
	\label{fig:radius}
	\vspace{-5mm}
\end{figure}

Here we introduce some notation for dealing with transformation matrices in $SE(3)$. A transformation $\mbf{T} \in SE(3)$ is related to its associated Lie algebra $\mbs{\xi}^{\wedge} \in \mathfrak{se}(3)$ through the exponential map:
\vspace{-1mm}
\begin{equation}
	\mbf{T} = \bbm \mbf{C} & \mbf{r} \\ \mbf{0}^T & 1 \ebm = \text{exp}(\mbs{\xi}^{\wedge}),
\end{equation}
where $\mbf{C}$ is $3 \times 3$ a rotation matrix, $\mbf{r}$ is a $3 \times 1$ translation vector, and $(\cdot)^{\wedge}$ is an overloaded operator that converts a vector of rotation angles $\mbs{\phi}$ into a member of $\mathfrak{so}(3)$ and $\mbs{\xi}$ into a member of $\mathfrak{se}(3)$:
\vspace{-1mm}
\begin{align}
		\mbs{\phi}^\wedge &= \bbm \phi_1 \\ \phi_2 \\ \phi_3 \ebm^\wedge := \bbm 0 & -\phi_3 & \phi_2 \\ \phi_3 & 0 & -\phi_1 \\ -\phi_2 & \phi_1 & 0 \ebm, \\
		\mbs{\xi}^\wedge &= \bbm \mbs{\rho} \\ \mbs{\phi} \ebm^\wedge := \bbm \mbs{\phi}^\wedge & \mbs{\rho} \\ \mbf{0}^T & 1 \ebm.
\end{align}
\vspace{-1mm}Given our constant-velocity assumption, we can convert from a velocity vector $\mbs{\varpi}$ into a transformation matrix using the following formula:
\begin{align}
	\mbf{T} = \text{exp}(\Delta t \mbs{\varpi}^{\wedge}). \label{tform}
\end{align}


\vspace{-1mm}
In order to optimize our objective function $J(\mbs{\varpi})$, we first need to derive the relationship between $\mbf{T}_m$ and $\mbs{\varpi}$. The velocity vector $\mbs{\varpi}$ can be written as the sum of a nominal velocity $\overline{\mbs{\varpi}}$ and a small perturbation $\delta \mbs{\varpi}$. This lets us rewrite the transformation $\mbf{T}_m$ as the product of a nominal transformation $\overline{\mbf{T}}_m$ and a small perturbation $\delta \mbf{T}_m$:
\begin{align}
	\mbf{T}_m &= \text{exp}(\Delta t_m (\overline{\mbs{\varpi}} + \delta \mbs{\varpi})^{\wedge}) = \delta \mbf{T}_m \overline{\mbf{T}}_m.
\end{align}
Let $\mbf{g}_m(\mbs{\varpi}) := \mbf{T}_m \mbf{p}_{m,1}$, which is nonlinear due to the transformation. Our goal is to linearize $\mbf{g}_m(\mbs{\varpi})$ about a nominal operating point. We can rewrite $\mbf{g}_m(\mbs{\varpi})$ as:
\begin{align}
	\mbf{g}_m(\mbs{\varpi}) &= \text{exp}(\Delta t_m \delta \mbs{\varpi}^\wedge) \overline{\mbf{T}}_m \mbf{p}_{m,1}, \nonumber \\
	&\approx (\mbf{1} + \Delta t_m \delta \mbs{\varpi}^\wedge)\overline{\mbf{T}}_m \mbf{p}_{m,1},
\end{align}
where we use an approximation for small pose changes. We swap the order of operations using the $(\cdot)^\pftm$ operator \cite{barfoot2017state}:
\begin{figure} [t]
	\centering
	
	\begin{tikzpicture} [axis/.style={>=latex,line width=1.5pt},
		vector/.style={line width=0.8pt},
		scale=1.20]
		
		\coordinate (I) at (0, 0);
		\draw [->, axis] (I) -- ($ (I) + (0, 3.5) $) node [midway, left] {$f$};
		\draw [->, axis] (I) -- ($ (I) + (6.0, 0) $) node [midway, below] {$t$};
		
		\def \x{2.5};
		\coordinate (A) at (0, 0.5);
		\coordinate (B) at ($ (A) + (\x, \x) $);
		\coordinate (C) at ($ (B) + (0, -\x) $);
		\coordinate (D) at ($ (C) + (\x, \x) $);
		\coordinate (E) at ($ (D) + (0, -\x) $);
		\coordinate (F) at ($ (E) + (1.0, 1.0) $);
		
		\draw [-, line width = 2.0pt, blue] (A) -- (B) -- (C) -- (D) -- (E) -- (F);
		
		\coordinate (A2) at (0.5, 0.65);
		\coordinate (O2) at ($ (A2) + (0, \x) $);
		\coordinate (P2) at ($ (O2) + (-0.5, -0.5) $);
		\coordinate (B2) at ($ (A2) + (\x, \x) $);
		\coordinate (C2) at ($ (B2) + (0, -\x) $);
		\coordinate (D2) at ($ (C2) + (\x, \x) $);
		\coordinate (E2) at ($ (D2) + (0, -\x) $);
		\coordinate (F2) at ($ (E2) + (0.5, 0.5) $);
		
		\draw [-, line width = 2.0pt, red] (P2) -- (O2) -- (A2) -- (B2) -- (C2) -- (D2) -- (E2) -- (F2);
		
		\coordinate (mid1) at ($ (A) + (0.4 * \x, 0.4 * \x) $);
		\coordinate (mid2) at ($ (mid1) + (0, -0.4) $);
		
		\draw [-, line width = 1.25pt] (mid1) -- (mid2);
		\draw [-, line width = 1.25pt] ($ (mid1) + (-0.25, 0) $) -- ($ (mid1) + (+0.25, 0) $);
		\draw [-, line width = 1.25pt] ($ (mid2) + (-0.25, 0) $) -- ($ (mid2) + (+0.25, 0) $);
		
		\draw node at ($ (mid1) + (0.0, 0.35) $) {$\Delta f$};
		
		\coordinate (mid3) at ($ (A) + (0.7 * \x, 0.7 * \x) $);
		\coordinate (mid4) at ($ (mid3) + (0.4, 0) $);
		\draw [-, line width = 1.25pt] (mid3) -- (mid4);
		\draw [-, line width = 1.25pt] ($ (mid3) + (0, -0.25) $) -- ($ (mid3) + (0, 0.25) $);
		\draw [-, line width = 1.25pt] ($ (mid4) + (0, -0.25) $) -- ($ (mid4) + (0, 0.25) $);
		
		\draw node at ($ (mid3) + (0.20, 0.55) $) {$\Delta t$};
		
		\coordinate (mid5) at ($ (C) + (\x / 2 - 0.25, \x / 2 - 0.25) $);
		\draw [-, line width = 1.25pt] (mid5) -- ($ (mid5) + (0, 0.5) $) -- ($ (mid5) + (0.5, 0.5) $);
		
		\draw node at ($ (mid5) + (0.25, 0.75) $) {$df / dt$};
		
		\draw [-, line width = 1.25pt] ($ (C) + (-0.25, 0) $) -- ($ (C) + (0.75, 0) $);
		\draw [-, line width = 1.25pt] ($ (C2) + (-0.75, 0) $) -- ($ (C2) + (0.25, 0) $);
		
		\draw node at ($ (C) + (1.0, 0) $) {$f_d$};
		
	\end{tikzpicture}
	\caption{This figure depicts the sawtooth modulation pattern of an FMCW radar. The transmitted signal is blue and the received signal is red.}
	\label{fig:sawtooth}
\end{figure}
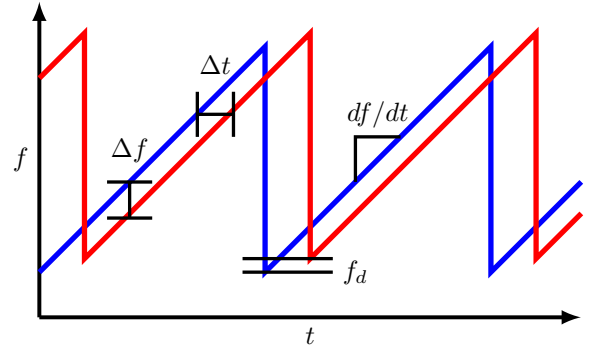
\begin{align}
	\mbf{g}_m(\mbs{\varpi}) &= \overline{\mbf{T}}_m \mbf{p}_{m,1} + \Delta t_m (\overline{\mbf{T}}_m \mbf{p}_{m,1})^\pftm \delta \mbs{\varpi} \nonumber \\
	&= \overline{\mbf{g}}_m + \mbf{G}_m \delta \mbs{\varpi}, \\
	\mbf{p}^\pftm &= \bbm \mbs{\rho} \\ \eta \ebm ^\pftm = \bbm \eta \mbf{1} & -\mbs{\rho}^\wedge \\ \mbf{0}^T & \mbf{0}^T \ebm.
\end{align}
We can now rewrite the error function from~\eqref{error1}:
\vspace{-1mm}
\begin{align}
	\mbf{e}_m &\approx \mbf{p}_{m,2} - \overline{\mbf{g}}_m - \mbf{G}_m \delta \mbs{\varpi} \nonumber \\
	&= \overline{\mbf{e}}_m - \mbf{G}_m \delta \mbs{\varpi}.
\end{align}

By inserting this equation for the error function into the objective function from~\eqref{objective}, and taking the derivative with respect to the perturbation and setting it to zero, $\frac{\partial J(\mbs{\varpi})} {\partial \delta \mbs{\varpi}^T} = 0$, we obtain the optimal update:
\begin{align}
	\delta \mbs{\varpi}^\star = \left(\sum_m \mbf{G}_m^T \mbf{R}_{{\rm cart},m}^{-1} \mbf{G}_m\right)^{-1} \left(\sum_m \mbf{G}_m^T \mbf{R}_{{\rm cart},m}^{-1} \overline{\mbf{e}}_m \right),
\end{align}
where $\mbf{R}_{{\rm cart},m} = \mbf{H}_m \mbf{R}_{m,2} \mbf{H}_m^T$ is the covariance in the local Cartesian frame, $ \mbf{h}(\cdot) = \mbf{f}^{-1}(\cdot)$, and $\mbf{H}_m = \left. \frac{\partial \mbf{h}}{\partial \mbf{x}} \right|_{\overline{\mbf{g}}_m}$. The optimal perturbation $\delta \mbs{\varpi}^\star$ is used in a Gauss-Newton optimization scheme and the process repeats until  $\mbs{\varpi}$ converges.


This method allows us to estimate the linear and angular velocity between a pair of radar scans directly while accounting for motion distortion. These velocity estimates can then be used to remove the motion distortion from a measurement relative to a reference time using~\eqref{tform}. MC-RANSAC is intended to be a lightweight method for showcasing the effects of motion distortion. A significant improvement to this pipeline would be to use the inliers of MC-RANSAC as an input to another estimator, such as \cite{anderson_iros15}. The inliers could also be used for mapping and localization or SLAM.




\subsection{Doppler Correction} \label{dopp_correct}

In order to compensate for Doppler effects, we need to know the linear velocity of the sensor $v$. This can either be obtained from a GPS/IMU or using one of the estimators described above. As shown in Figure~\ref{fig:radius}, the motion of the sensor results in an apparent relative velocity between the sensor and its surrounding environment. This relative velocity causes the received frequency to be altered according to the Doppler effect. Note that only the radial component of the velocity \change{$u = v_x \cos(\phi) + v_y \sin(\phi)$} will result in a Doppler shift. The Radar Handbook by Skolnik \cite{skolnik1990radar} provides an expression for the Doppler frequency:
\begin{equation}
		f_d = \frac{2 u}{\lambda}, \label{doppler}
\end{equation}
where $\lambda$ is the wavelength of the signal. Note that for an object moving towards the radar ($u > 0$) or vice versa, the Doppler frequency will be positive resulting in a higher received frequency. For FMCW radar such as the Navtech sensor, the distance to a target is determined by measuring the change in frequency between the received signal and the carrier wave $\Delta f$:
\begin{equation} \label{range}
	r = \frac{c \Delta f}{2 (df / dt)}, 
\end{equation}
where $df / dt$ is the slope of the modulation pattern used by the carrier wave and $c$ is the speed of light. FMCW radar require two measurements to disentangle the frequency shift resulting from range and relative velocity. Since the Navtech sensor scans each azimuth only once, the measured frequency shift is the combination of both the range difference and Doppler frequency. From Figure~\ref{fig:sawtooth}, we can see that a positive Doppler frequency $f_d$ will result in an increase in the received frequency and in turn a reduction in the observed frequency difference $\Delta f$. Thus, a positive Doppler frequency will decrease the apparent range of a target.


The Navtech radar operates between 76 GHz and 77~GHz resulting in a bandwidth of 1 GHz. Navtech states that they use a sawtooth modulation pattern. Given 1600 measurements per second and assuming the entire bandwidth is used for each measurement, $df / dt \approx 1.6 \times 10^{12}$.

Hence, if the forward velocity of the sensor is 1 m/s, a target positioned along the $x$-axis (forward) of the sensor would experience a Doppler frequency shift of 510 Hz using~\eqref{doppler}. This increase in the frequency of the received signal would decrease the apparent range to the target by 4.8 cm using~\eqref{range}. Naturally, this effect becomes more pronounced as the velocity increases.


Let $\beta = f_t / (df / dt)$ where $f_t$ is the transmission frequency $(f_t \approx 76.5 \mbox{ GHz})$. In order to correct for the Doppler distortion, the range of each target needs to be corrected by the following factor:
\begin{equation}
	\Delta r_{\rm corr} = \beta (v_x \cos(\phi) + v_y \sin(\phi)).
\end{equation}
We use this simple correction in all our experiments with the velocity \change{$(v_x, v_y)$} coming from our motion estimator.


\section{Experimental Results}

In order to answer the question posed by this paper, we have devised two experiments. The first is to compare the performance of rigid RANSAC and MC-RANSAC (with or without Doppler corrections) on radar odometry using the Oxford Radar RobotCar Dataset \cite{oxford_radar}. The second \change{experiment demonstrates the impact of both distortion effects on} localization using our own data. We also provide \change{qualitative results demonstrating both distortion effects.}




The Navtech radar is a frequency modulated continuous wave (FMCW) radar. For each azimuth, the sensor outputs the received signal power at each range bin. The sensor spins at 4 Hz and provides 400 measurements per rotation with a 163 m range, 4.32 cm range resolution, 1.8$^\circ$ horizontal beamwidth, and 0.9$^\circ$ azimuth resolution. 


The Oxford Radar RobotCar Dataset is an autonomous driving dataset that includes two 32-beam lidars, six cameras, a GPS/IMU, and the Navtech sensor. The dataset includes thirty-two traversals equating to 280 km of driving in total. 


\vspace{-1mm}

\subsection{Odometry} \label{sec:odom}

\begin{table}[t]
	\caption{Radar Odometry Results. *Odometry results from \cite{barnes_icra20}.}
	\begin{tabular}{|l|c|c|}
		\hline
		Method                                                       & Translational Error (\%) & Rotational Error (deg/m) \\ \hline
		RO Cen* \cite{cen_icra18}                                                      & 3.7168                   & 0.0095                   \\ \hline
		Rigid RANSAC                                                 & 3.9777                   & 0.0141                   \\ \hline
		MC-RANSAC                                                     & 3.6042                   & 0.0119                   \\ \hline
		\begin{tabular}[c]{@{}l@{}}MC-RANSAC +\\ Doppler\end{tabular} & 3.5847                   & 0.0118                   \\ \hline
	\end{tabular}
	
	\label{tab:odom}
	\vspace{-2mm}
\end{table}

Our goal is to make a fair comparison between two estimators where the main difference is the compensation of distortion effects. To do this, we use the same number of maximum iterations (100), and the same inlier threshold (0.35 m) for both rigid RANSAC and MC-RANSAC. We also fix the random seed before running either estimator to ensure that the differences in performance are not due to the random selection of subsets.

For feature extraction, we use the same setup parameters for Cen2018 as in \cite{cen_icra18} except that we use a higher probability threshold, $z_q = 3.0$, and a Gaussian filter with $\sigma = 17$ range bins. We convert polar radar scans into a Cartesian image with 0.2592 m per pixel and a width of 964 pixels (250 m). For ORB descriptors, we use a patch size of 21 pixels (5.4 m). For data association, we use a nearest-neighbor distance ratio of 0.8. For Doppler corrections, we use $\beta = 0.049$. For each radar scan, extracting Cen2018 features takes approximately 35 ms. Rigid RANSAC runs in 1-2 ms and MC-RANSAC in 20-50 ms. Calculating orb descriptors takes 5 ms and brute-force matching takes 15 ms. \change{These processing times were obtained on a quad-core Intel Xeon E3-1505M 3.0 GHz CPU with 32 GB of RAM.}

Odometry results are obtained by compounding the frame-to-frame scan matching results. Note that we do not use a motion prior or perform additional smoothing on the odometry. Three sequences were used for parameter tuning. The remaining 29 sequences are used to provide test results.


Table~\ref{tab:odom} summarizes the results of the odometry experiment. We use KITTI-style odometry metrics \cite{geiger_ijrr13} to quantify the translational and rotational drift as is done in \cite{barnes_icra20}. The metrics are obtained by averaging the translational and rotational drifts for all subsequences of lengths (100, 200, ..., 800) meters. The table shows that motion-compensated RANSAC results in a \change{9.4\%} reduction in translational drift and a 15.6\% reduction in rotational drift. This shows that compensating for motion distortion has a modest impact on radar odometry. The table also indicates that Doppler effects have a negligible impact on odometry. Our interpretation is that Doppler distortion impacts sequential radar scans similarly such that scan registration is minimally impacted.


It should be noted that a large fraction of the Oxford dataset was collected at low speeds (0-5 m/s). This causes the motion distortion and Doppler effects to be less noticeable.



\begin{figure} [t]
	\centering
	\includegraphics[width=\columnwidth]{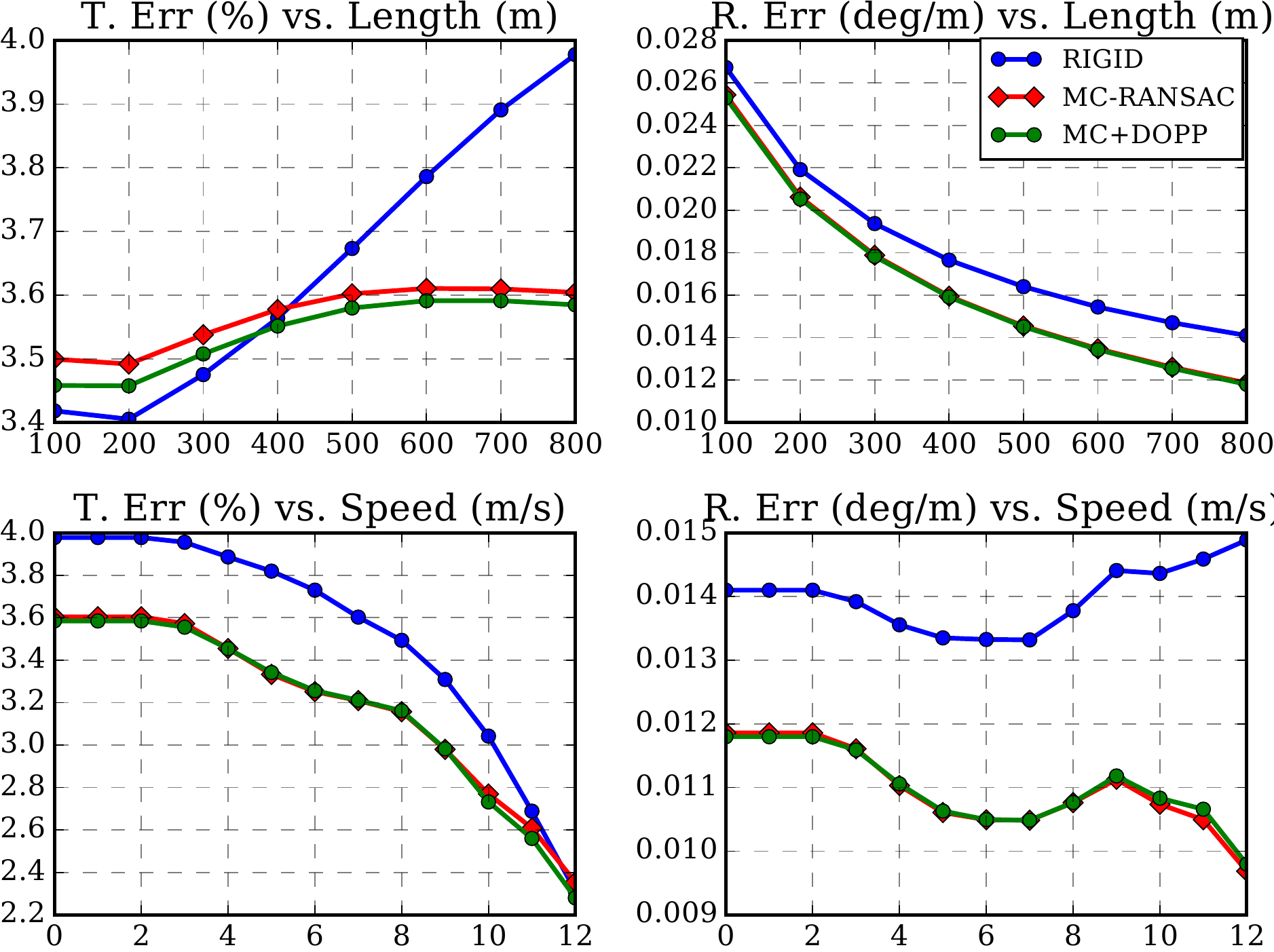}
	\caption{This figure provides our KITTI-style odometry results on the Oxford dataset. We provide our translational and rotational drift as a function of path length and speed. MC-RANSAC: motion-compensated RANSAC, MC+DOPP: motion and distortion compensated.}
	\label{fig:odometry}
	\vspace{-5mm}
\end{figure}

\begin{figure} [t]
	\centering
	\includegraphics[width=\columnwidth]{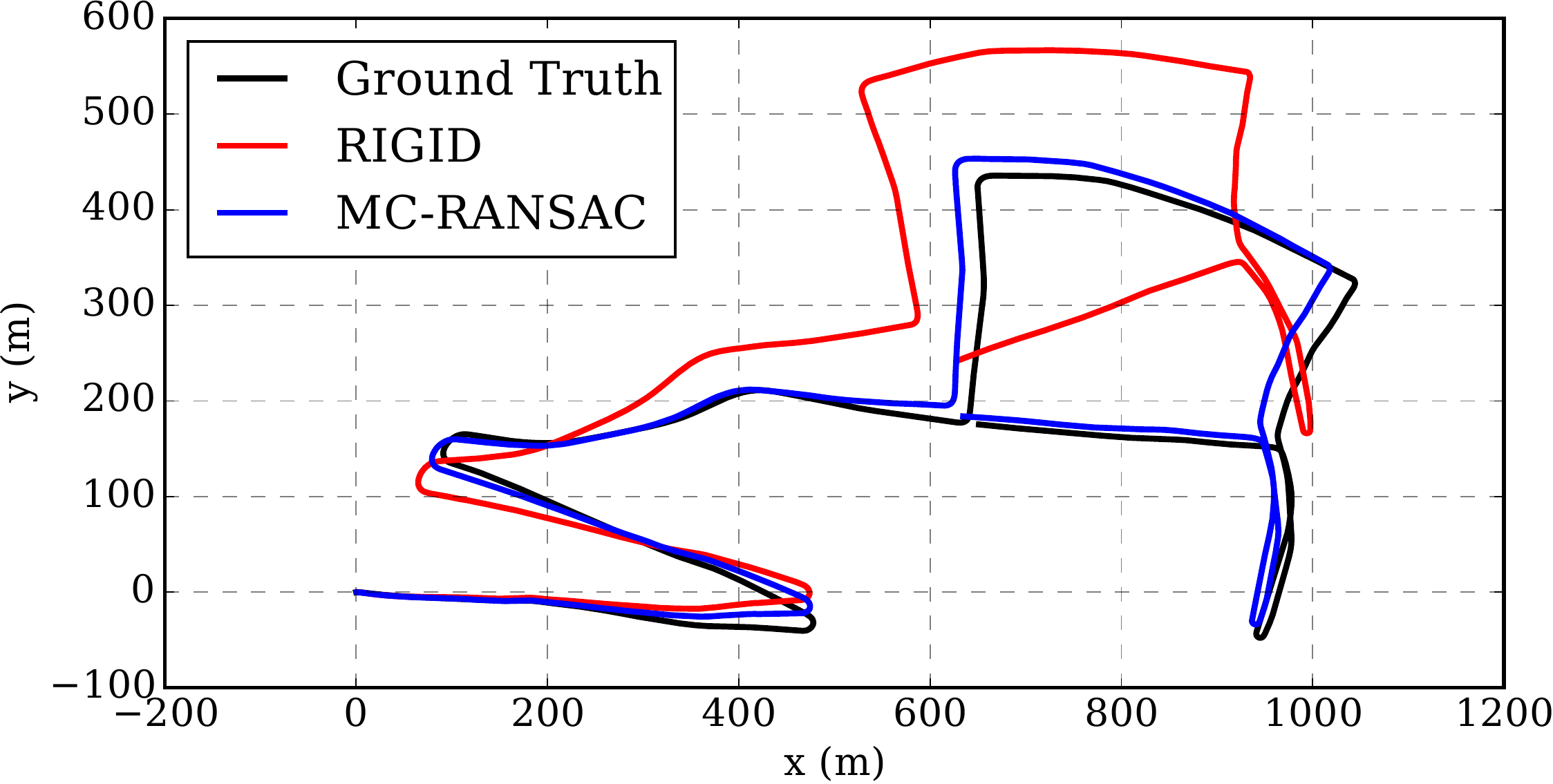}
	\caption{This figure highlights the impact that motion distortion can have on the accuracy of radar-based odometry. Note that motion-compensated RANSAC (MC-RANSAC) is much closer to the ground truth.}
	\label{fig:trajectory}
\end{figure}

Figure~\ref{fig:odometry} depicts the translational and rotational drift of each method as a function of path length and speed. It is clear that motion compensation improves the odometry performance across \change{most} cases. Interestingly, MC-RANSAC does not increase in error as much as rigid RANSAC as the path length increases. Naturally, we would expect rigid RANSAC to become much worse than MC-RANSAC at higher speeds. However, what we observe is that as the speed of the vehicle increases, the motion tends to become more linear. When the motion of the vehicle is mostly linear, the motion distortion does not impact rigid RANSAC as much. Figure~\ref{fig:trajectory} compares the odometry output of both estimators against ground truth. The results further illustrate that compensating for motion distortion improves performance.



\vspace{-1mm}

\subsection{Localization}

The purpose of this experiment is to demonstrate the impact of motion distortion and Doppler effects on metric localization. As opposed to the previous experiment, we localize between scans taken while driving in opposite directions. While the majority of the Oxford Radar dataset was captured at low speeds (0-10 m/s), in this experiment we only use radar frames where the ego-vehicle's speed was above 10 m/s. For this experiment, we use our own data-taking platform, shown in Figure~\ref{fig:buick}, which includes a Velodyne Alpha-Prime lidar, Navtech CIR204-H radar, Blackfly S camera, and an Applanix POSLV GNSS. \change{Individuals interested in using this data can fill out a Google form to gain access \footnote{\scriptsize{\url{https://forms.gle/ZGtQhKRXkxmcAGih9}}}}.

Ground truth for this experiment was obtained from a 10 km drive using post-processed GNSS data provided by Applanix, which has an accuracy of 12 cm in this case. Radar scans were initially matched by identifying pairs of proximal scans on the outgoing and return trips based on GPS data. The Navtech timestamps were synchronized to GPS time to obtain an accurate position estimate.

Our first observation was that localizing against a drive in reverse is harder than odometry. When viewed from different angles, objects have different radar cross sections, which causes them to appear differently. As a consequence, radar scans may lose or gain features when pointed in the opposite direction. This change in the radar scan's appearance was sufficient to prevent ORB features from matching. 


As a replacement for ORB descriptors, we turned to the Radial Statistics Descriptor (RSD) described in \cite{cen_icra18}, \cite{cen_icra19}. Instead of calculating descriptors based on the Cartesian radar image, RSD operates on a binary Cartesian grid derived from the detected feature locations. This grid can be thought of as a radar target occupancy grid. For each keypoint, RSD divides the binary grid into M azimuth slices and N range bins centered around the keypoint. The number of keypoints (pixels) in each azimuth slice and range bin is counted to create two histograms. In \cite{cen_icra19}, a fast Fourier transform of the azimuth histogram is concatenated with a normalized range histogram to form the final descriptor.


In our experiment, we found that the range histogram was sufficient on its own, with the azimuth histogram offering only a minor improvement. It should be noted that these descriptors are more expensive to compute (60 ms) and match (30 ms) than ORB descriptors. \change{These processing times were obtained using the same hardware as in Section~\ref{sec:odom}}.



\begin{figure} [t]
	\centering
	\includegraphics[width=\columnwidth]{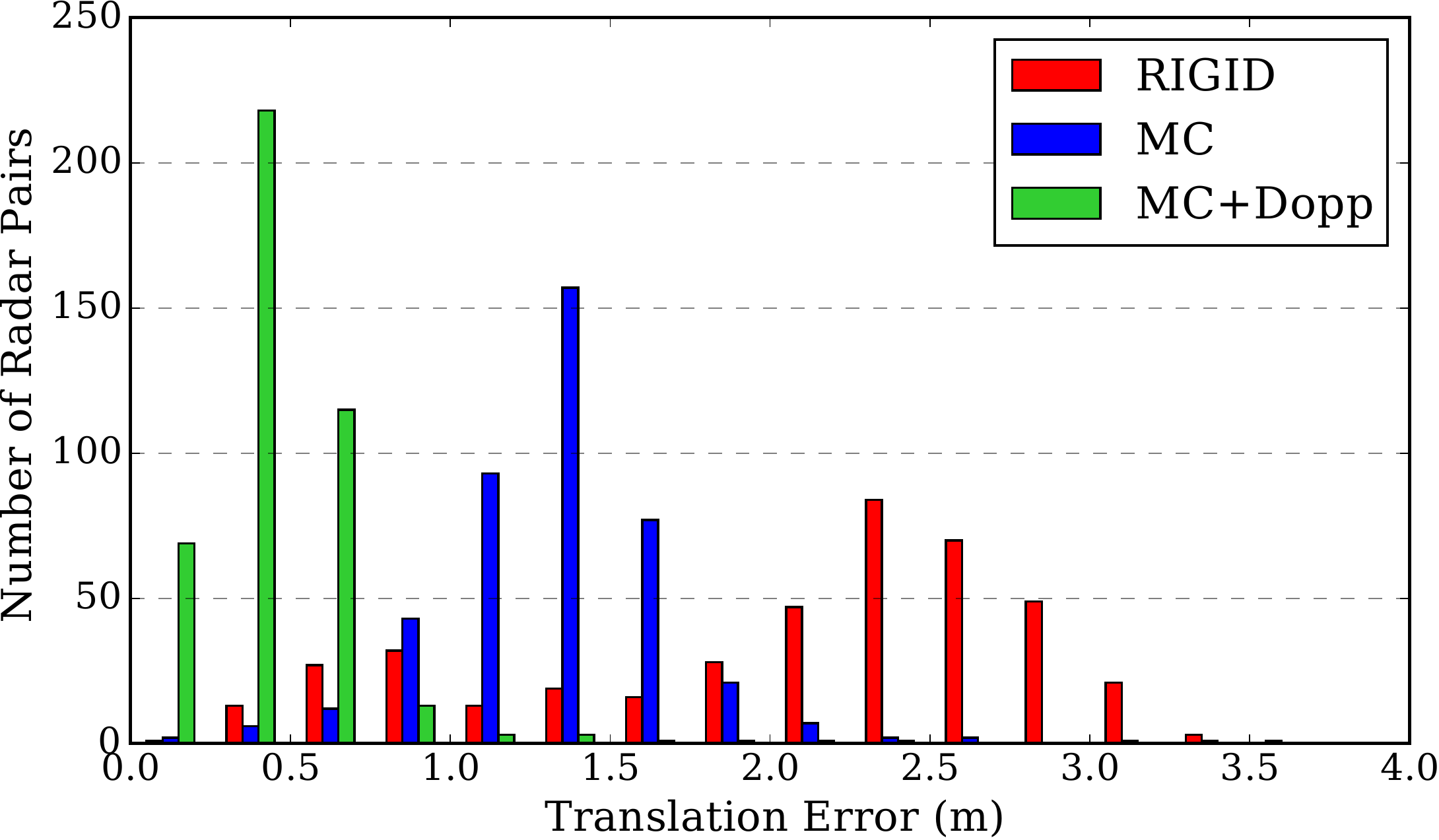}
	\caption{By compensating for motion distortion alone (MC) or both motion and Doppler distortion (MC +  DOPP), metric localization improves.}
	\label{fig:localization}
	\vspace{-2mm}
\end{figure}

The results of our localization experiment are summarized in Table~\ref{tab:loc}. In each case, we are using our RANSAC estimator from Section~\ref{sec:method}. The results in the table are obtained by calculating the median translation and rotation error. Compensating for motion distortion results in a \change{41.7\%} reduction in translation error. Compensating for Doppler effects results in a further \change{67.7\%} reduction in translation error. Together, compensating for both effects results in a \change{81.2\%} reduction in translation error. Note that the scan-to-scan translation error is larger than in the odometry experiment due to the increased difficulty of localizing against a reverse drive. Figure~\ref{fig:localization} depicts a histogram of the localization errors in this experiment.

\vspace{-1mm}



\begin{table}[h]
	\centering
	\caption{Metric Localization Results}
	\begin{tabular}{|l|c|c|}
		\hline
		Method                                                       & Trans. Error (m) & Rot. Error (deg) \\ \hline
		No Compensation                                                 & 2.2976                  & 0.7110                   \\ \hline
		Motion-Compensated                                                     & 1.3403                  & 1.0766                   \\ \hline
		\begin{tabular}[c]{@{}l@{}}Motion-Compensated +\\ Doppler-Compensated \end{tabular} & 0.4327                  & 0.9984                   \\ \hline
	\end{tabular}
	\label{tab:loc}
\end{table}

\newpage
\subsection{\change{Qualitative Results}}



\begin{figure}[h]
	\vspace{-0.5mm}
	\centering
	\subfigure[Distorted]{\includegraphics[width=0.85\columnwidth]{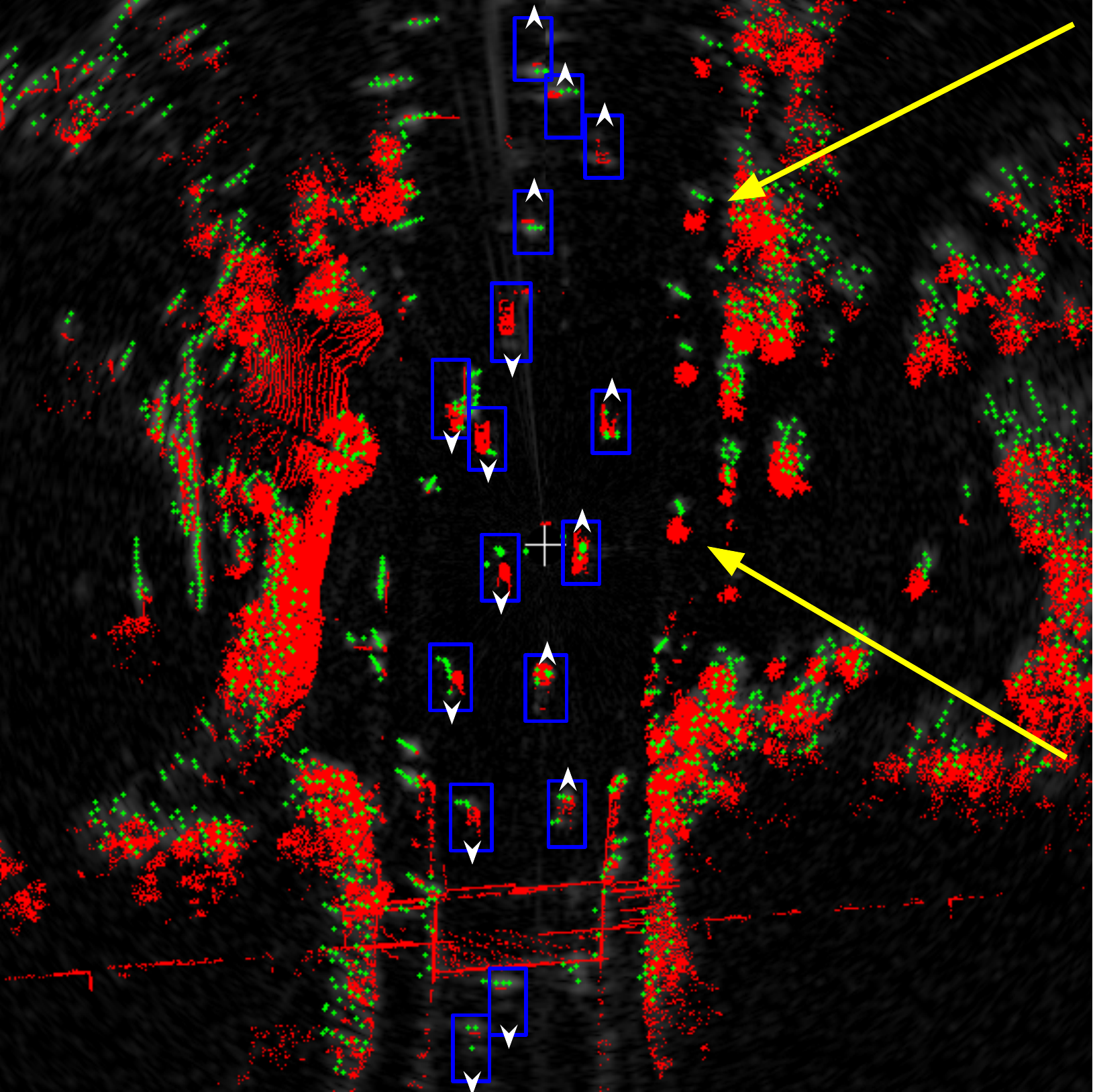}}
	\subfigure[Motion and Doppler Distortion Removed]{\includegraphics[width=0.85\columnwidth]{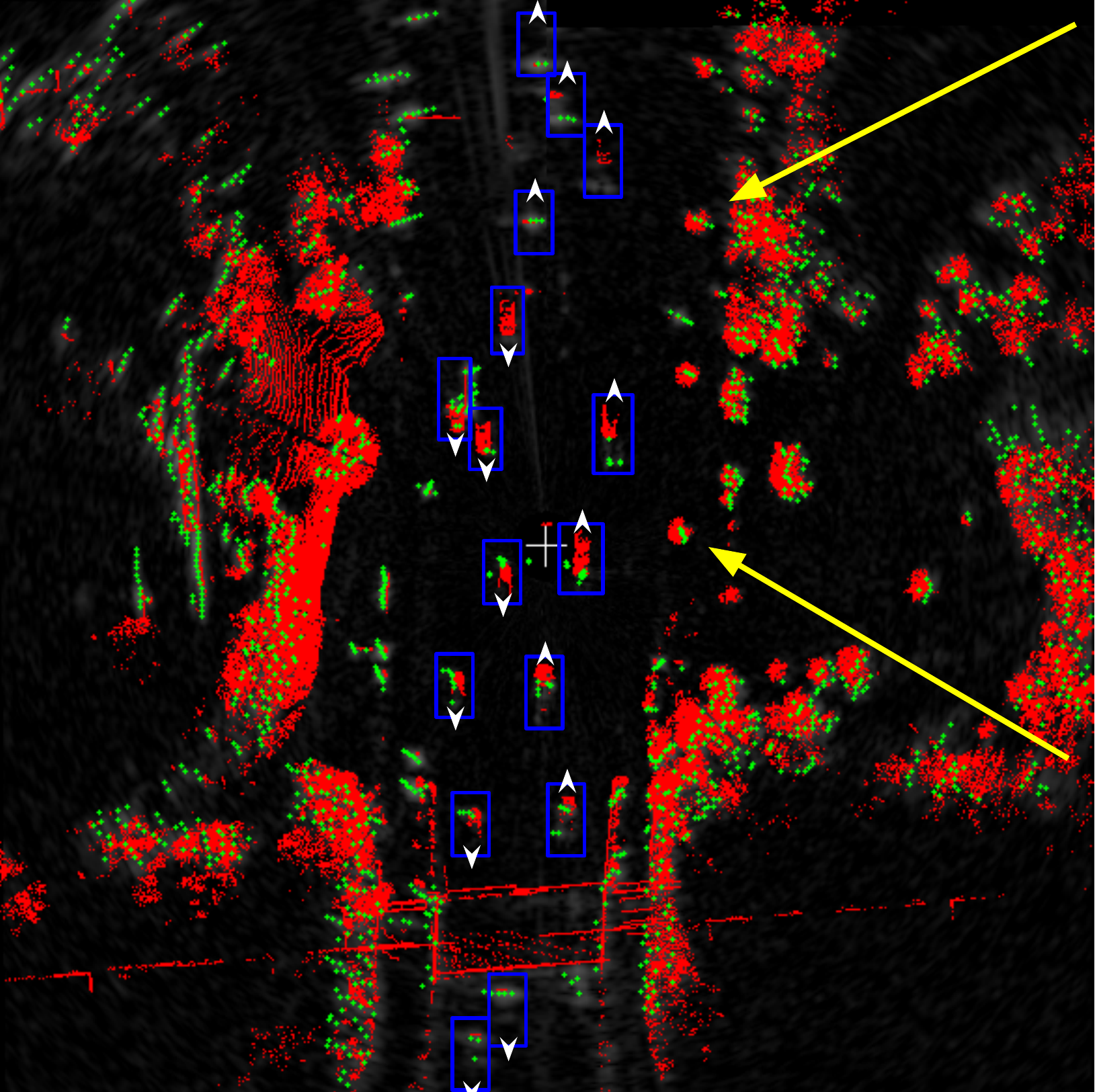}}
	\caption{\change{Lidar points are shown in red, radar targets are shown in green, vehicles are boxed in blue. (a) Both motion distortion and Doppler effects and present in the radar scan. (b) Motion distortion and Doppler effects have been removed from the radar data. Note that static objects (highlighted by yellow arrows) align much better in (b). Some of the moving vehicles (boxed in blue) are less aligned after removing distortion. Here we do not know the other vehicles' velocities and therefore the true relative velocity with respect to the ego-vehicle.}}
	\label{fig:distortion}
\end{figure}

\change{In this section, we present qualitative results of removing motion distortion and Doppler effects from radar data.} In Figure~\ref{fig:distortion}, we have plotted points from our Velodyne lidar in red \change{and extracted radar targets in green.} In this example, the ego-vehicle is driving at \change{15 m/s} with vehicles approaching from the opposite direction on the left-hand side of the road. \change{In order to directly compare lidar and radar measurements, we have aligned each sensor's output spatially and temporally using post-processed GPS data and timestamps. Motion distortion is removed from the lidar points before the comparison is made. We use the lidar measurements to visualize the amount of distortion present in the radar scan. } Figure~\ref{fig:distortion}(a) shows what the original alignment looks like when \change{the radar scan is} distorted. Figure~\ref{fig:distortion}(b) shows what the alignment looks like after compensating for motion distortion and Doppler effects. Note that static objects such as trees align much better with the lidar data in Figure~\ref{fig:distortion}(b). \change{It interesting to note that some of the moving vehicles (boxed in blue) are less aligned after removing distortion. Here we do not know the other vehicles' velocities and therefore the true relative velocity with respect to the ego-vehicle. These results indicate that additional velocity information for each object is required in order to align dynamic objects. We leave this problem for future work.}




\vspace{-1.0mm}
\section{Conclusion}
\vspace{-1.0mm}



For the problem of odometry, compensating for motion distortion had a \change{modest} impact of reducing translational drift by \change{9.4\%}. Compensating for Doppler effects had a negligible effect on odometry performance. We postulate that Doppler effects are negligible for odometry because their effects are quite similar from one frame to the next. In our localization experiment, we observed that compensating for motion distortion and Doppler effects reduced translation error by \change{41.7\%} and \change{67.7\%} respectively with a combined reduction of \change{81.2\%}. \change{We also provided qualitative results demonstrating the impact of both distortion effects. We noted that the velocity of each dynamic object is required in order to properly undistort points associated with dynamic objects.} In summary, the Doppler effect can be safely ignored for the radar odometry problem but motion distortion should be accounted for to achieve the best results. For metric localization, especially for localizing in the opposite direction from which the map was built, both motion distortion and Doppler effects \change{need to} be compensated. Accounting for these effects is computationally cheap, but requires an accurate estimate of the linear and angular velocity of the sensor.


For future work, we will investigate applying more powerful estimators such as \cite{anderson_iros15} to odometry and the full mapping and localization problem. We will also investigate learned features and the impact of seasonal changes on radar maps.

\vspace{-1mm}


\section*{Acknowledgement}

We would like to thank Goran Basic for designing and assembling our roof rack for Boreas. We would also like to thank Andrew Lambert and Keith Leung at Applanix for their help in integrating the POSLV system and providing the post-processed GNSS data. We thank General Motors for their donation of the Buick.




\bibliography{IEEEabrv,bib/references}

\end{document}